\documentclass[letterpaper, 10 pt, conference]{ieeeconf}  %

\IEEEoverridecommandlockouts                              %

\title{\LARGE \bf Simulation of Vision-based Tactile Sensors\\ using Physics based Rendering}

\author{Arpit Agarwal$^{1}$, Timothy Man$^{2}$ and Wenzhen Yuan$^{1}$%
\thanks{$^{1}$Arpit Agarwal and Wenzhen Yuan are with the Robotics Institute, Carnegie Mellon University, 5000 Forbes Ave, Pittsburgh, PA 15213, USA
        {\tt\small \{arpita1\}@andrew.cmu.edu, yuanwz@cmu.edu}}%
\thanks{$^{2}$Timothy Man is with the Department of Mechanical Engineering, Carnegie Mellon University, 5000 Forbes Ave, Pittsburgh, PA 15213, USA
        {\tt\small \{tman2\}@andrew.cmu.edu}}%
}

\usepackage[pdftex]{graphicx}
\graphicspath{{pdf/}{pngs/}}
\usepackage{cite}
\usepackage{amsmath}
\usepackage{algorithmic}
\usepackage{array}
\usepackage{url}
\usepackage{hyperref}
\usepackage{todonotes}
\hypersetup{
colorlinks=true, 
linkcolor=red
}

\usepackage{geometry}
\begin{document}
\maketitle

\newcommand{\wan}[1]{\textcolor{green}{#1}}
\newcommand{\an}[1]{\textcolor{red}{#1}}
\newcommand{\high}[1]{\colorbox{BurntOrange}{#1}}

\newcommand{\Wtext}[1]{\textcolor{blue}{#1}}

\begin{abstract}
Tactile sensing has seen a rapid adoption with the advent of vision-based tactile sensors. Vision-based tactile sensors provide high resolution, compact and inexpensive data to perform precise in-hand manipulation and human-robot interaction. However, the simulation of tactile sensors is still a challenge. In this paper, we built the first fully general optical tactile simulation system for a GelSight sensor using physics based rendering techniques. We propose physically accurate light models and show in-depth analysis of individual components of our simulation pipeline. Our system outperforms previous simulation techniques qualitatively and quantitative on image similarity metrics. Our code and experimental data is open-sourced at \href{https://labs.ri.cmu.edu/robotouch/tactile-optical-simulation/}{project page}.
\end{abstract}


\section{Introduction}

Simulation is a critical tool in the development of robotic systems. It is widely used for hardware design, control and planning. Simulations are useful not only at the start of the development process, but also for debugging and rapid iteration on the design when a new design objective emerges. Due to the above advantages, we have seen development of a number of rigid body simulators like ODE\cite{smith2005open}, SimBody\cite{sherman2011simbody}, MuJoCo\cite{todorov2012mujoco}, Dart\cite{lee2018dart} and, particle-based simulators like Nvidia FleX\cite{macklin2019non} and SOFA\cite{allard2007sofa}.
Tactile sensing is a cornerstone for complex robotic manipulation together with advanced control algorithms and hardware design. However, most modern simulators have limited tactile sensing simulation.

\begin{figure}
    \centering
    \includegraphics[width=0.7\columnwidth]{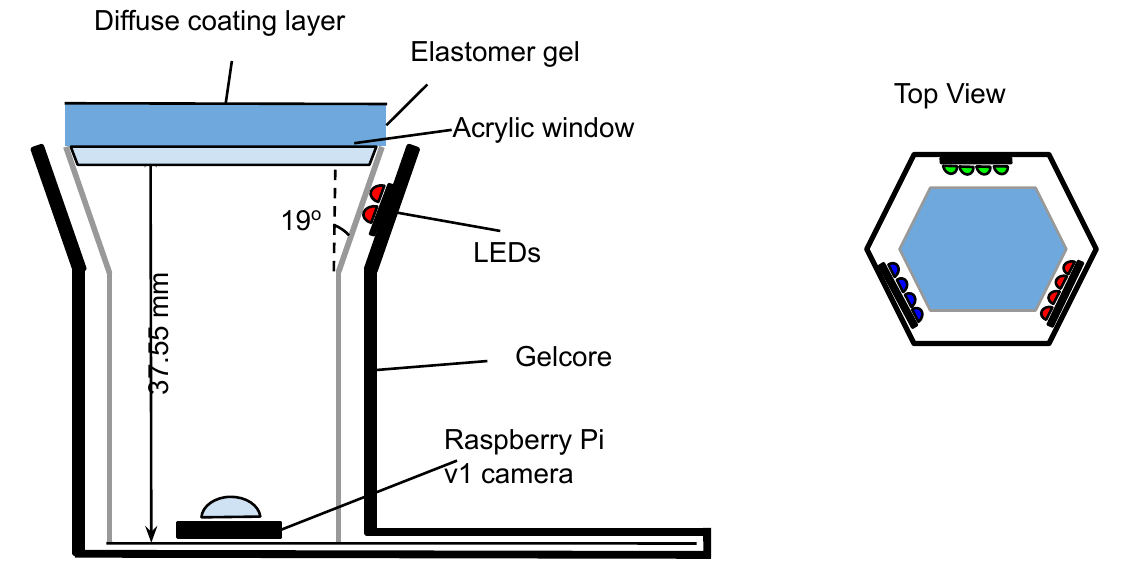}
    \caption{GelSight sensor illustration: The key components which we model in our project are gelcore, elastomer surface and LEDs.}
    \label{fig:illustration}
\end{figure}

Vision-based tactile sensors like GelSight\cite{gsSensors}, which we use in this paper, have multiple colored lights, soft deformable opaque skin and a RGB camera. Figure \ref{fig:illustration} shows the illustration of the GelSight prototype used in our study. The prototype is based on sensor proposed in \cite{gsSensors}. When an object interacts with the soft sensor skin, the deformed skin shape reflects light, probably multiple times, to form an image in the camera. The sensor uses photometric stereo \cite{basri2007photometric} to invert the RGB color information to shape information. We can obtain high resolution shape, multi-axis force and friction information using GelSight. 

Vision-based tactile sensor simulation have two major components, namely \textit{optical simulation} and \textit{dynamics simulation}. In this paper, we tackle \textit{optical simulation}, which refers to simulating interaction of light with the deformed skin and its subsequent image formation process by the camera. Our simulated images contain features like correct shadows, interreflections, light variation across the sensing surface, etc. which can be used to accurately sense geometry of object interacting with tactile sensors.
\begin{figure}[!t]
\centering
\includegraphics[keepaspectratio,width=\columnwidth]{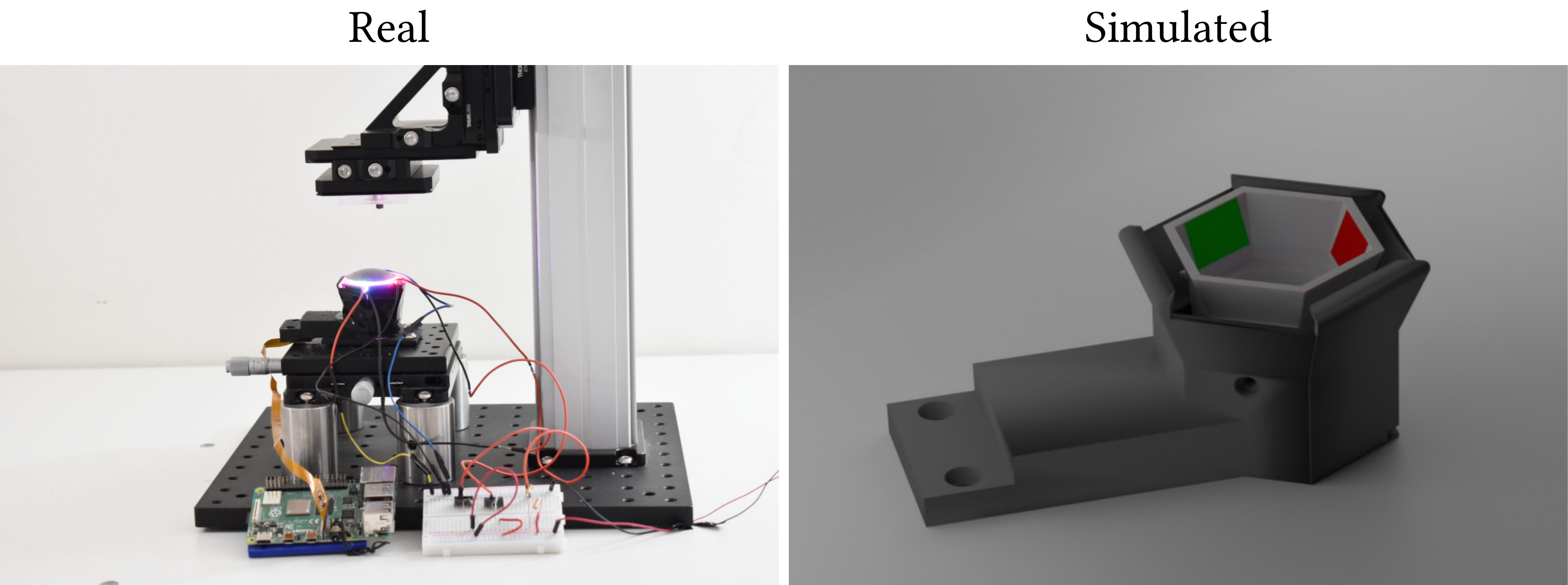}
\includegraphics[keepaspectratio,width=\columnwidth]{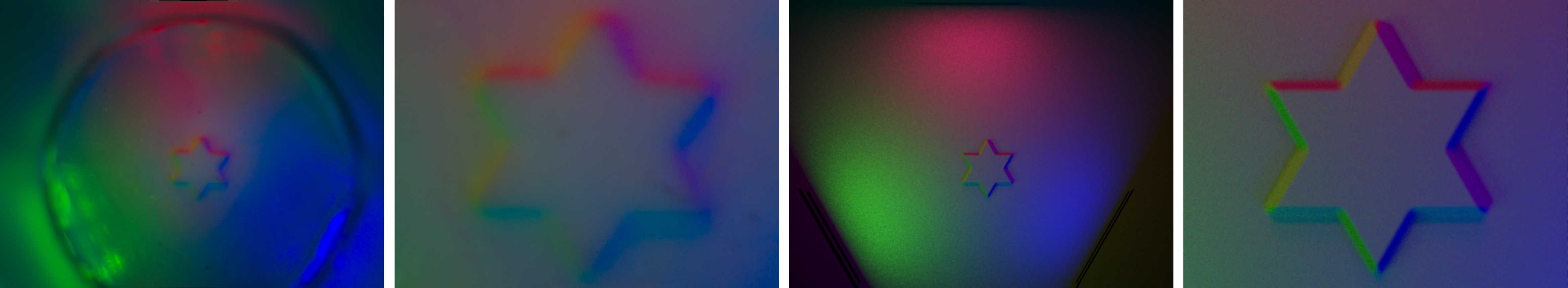}
\caption{The comparison of a real GelSight sensor and a simulated GelSight sensor, when a star-shaped 3D-printed object is contacting the sensor. Our model well simulated the optical system in the sensor, and therefore can generate a realistic tactile image that indicates the object's shape.}
\label{teaser}
\end{figure}

Initial attempts at simulating high-resolution vision-based  tactile sensors such as \cite{gomes2019gelsightsim} and \cite{roberto_shaperecon} make  use of ray casting techniques and non-physical models of lighting. These assumptions limit the use of these techniques only for simulating a specific sensor after calibration. 

In this work, we develop an optical simulation system using physics-based rendering(PBR)\cite{pharr2016physically} techniques. PBR focuses on accurately modeling physics of light scattering. PBR allows modifying the physical location of optical elements like cameras and lights; changing the deformable surface geometry; changing the optical properties of sensor surface(shiny vs diffuse). Therefore, our system can be used as a design tool and simulation software for producing accurate tactile images. Our system is able to successfully capture the variation in color, light intensity and shape at different location of the elastomer surface.

The physical models proposed in this paper can serve as a starting point to model other high resolution vision-based sensors and help in adapting the sensor to other form factors suitable for robotics. 
Our simulation model can help in the development of complex data-driven models and learn policies for complex robotic tasks.

\section{Related Work}

\textit{Contact simulation for tactile sensors}.
Previous approaches for modeling and simulating the contact between sensor surface with the object can be categorized into a) modeling the deformation of the tactile sensor surface, and b) modeling low-dimensional features used into a particular sensing technology.  
\cite{narang2020interpreting} simulated BioTac\cite{fishel2012sensing}, which is a finger-shaped sensor with a fluid-coupled electrode array and  measures impedances. In their simulation, BioTac was modeled using Finite Element Method(FEM) which outputs the quasi-static nodal displacement under applied force over a known contact area. 
Vision-based tactile sensors like TacTip\cite{ward2018tactip} and \cite{sferrazza2019design}, which track the motion of dots, either on sensor surface or embedded in fluid, have also been simulated. 
\cite{ding2020sim} modeled TacTip as a set of \textit{pin} positions(similar to found on the real sensor) which deform using elastic pushing and pulling force based on contact with objects in the scene. 
\cite{sferrazza2020learning} simulated their sensor optical-flow based tactile sensor\cite{sferrazza2019design} using Neural Network, trained using a dataset generated by Finite Element(FEM) simulation. Their simulation is able to predict the position and force distribution. They decoupled camera parameters from Neural Network training to allow adaptation to various cameras. This system could be used for simulating the surface deformations and low-dimensional tactile features. However, the above methods do not work well for GelSight due to complex light system.
\cite{zhang2016triangle} simulated the tactile sensor by  finding the intersection surface between the object mesh and the robot hand; and sampled points along that surface to simulate tactile sensation for force closure.

\textit{Tactile optical simulation}. There have been some recent works which simulate the image formation process for Gelsight like vision-based tactile sensors. \cite{roberto_shaperecon} and \cite{gomes2019gelsightsim} used directional lights, with phong material and diffuse material respectively, to simulate image formed by the tactile sensors. The assumption of directional light breaks down if the physical lights are very close to the scene(sensor surface in this case)\cite{liu2018near}. Our work uses a general image formation process which takes multiple bounces of light into account, together with physically accurate light model and surface material model. This allows us to capture the spatial variation in color and intensity distribution in the simulated image.

\section{Method}

A GelSight tactile sensor generates an image which is formed by the interaction of lights, emitted by LEDs, passing through a translucent supporting structure inside the sensor, then hitting the deformed soft elastomer and reaching the camera. The process of image formation using physics is known as light transport and is well studied under the field of physics-based rendering(PBR)\cite{pharr2016physically}. These techniques are heavily used in making movies\cite{pathtracingfilms} and designing optical elements\cite{nimier2019mitsuba}. Therefore, to simulate our tactile sensor, we use PBR. 

The section first provides (\ref{pathtracing}) an overview of physics-based rendering and then describes the (\ref{light_model})specific models of light, (\ref{gelcore_model})material of the translucent supporting structure (which is referred to as \textit{gelcore} in the rest of the paper), and (\ref{elastomer})elastomer used to create the GelSight sensor in simulation. 
\subsection{Light Transport Simulation}\label{pathtracing}
The process of tracking the path of light from LEDs to camera is known as light transport in PBR. However, a lot of the light rays do not reach the camera. Therefore, to reduce the computation cost, light path tracking begins at the camera and attempts to reach the light source. This technique is known as \textit{path tracing} in computer graphic, which forms the core of our simulation pipeline. 

\begin{figure}[!t]
\centering
\includegraphics[width=0.8\columnwidth]{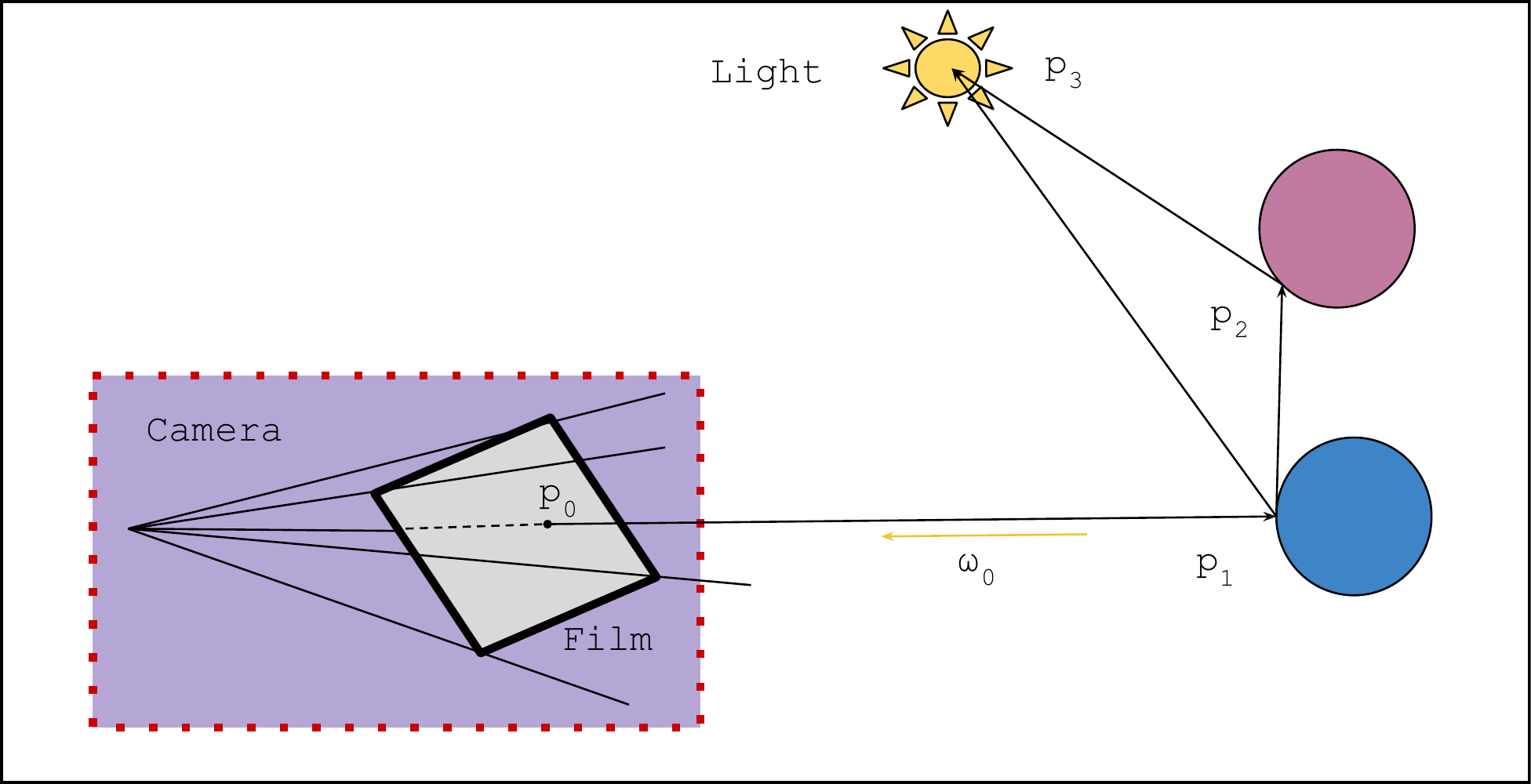}
\caption{\textit{Path Tracing}: image construction process, in terms of light paths, starts from camera, hits single or multiple objects and reaches the light source.}
\label{raytracing}
\end{figure}

Consider the scene shown in Figure \ref{raytracing}. The light radiance received at point $p_0$ from point $p_1$ along the direction $\omega_0$ is 
\begin{align*}
    L(p_1,\omega_0) = L_e(p_1,\omega_0) + \int_{S^2} &f(p_1,\omega_0, \omega_i) \\
    &L(t(p_1, \omega_i),-\omega_i)|cos \theta_i|d\omega_i
\end{align*}
where $L_e(p_1,\omega_0)$ is light emitted by point $p_1$ towards direction $\omega_0$; $f(p_1,\omega_0, \omega_i)$ is the material model, which gives the fraction of light emitted in direction $\omega_0$, when receiving light from direction $\omega_i$; $t(p, \omega_i)$ is the point in the light path visited prior to hitting point $p_1$ at an angle $\omega_i$; $|cos \theta_i|$ is the foreshortening term. On a high level, the first term represents light emitted at $p_1$ and the second term represents light emitted by all the points in the scene towards $p_1$ sampled according to some probability distribution. The rendering equation can not be solved analytically as it is a recursive equation(as term $L(p, \omega)$ appears on both sides of the equation) in high dimension for a generic scene. To solve it, path integral formulation of the above equation is considered, as shown below
\begin{align}
L&(p_1 \rightarrow p_0) = L_e(p_1 \rightarrow p_0) \\
        & + \int_A L(p_2 \rightarrow p_1)h(p_2 \rightarrow p_1 \rightarrow p_0)dA(p_2)\\
        & + \int_A\int_A L(p_3 \rightarrow p_2)h(p_3 \rightarrow p_2 \rightarrow p_1)dA(p_3)dA(p_2) \\
        &+ \ldots
\end{align}
According to the above equation, we need to generate all the paths starting from light sources, hitting different objects in the scene and reaching the camera. In practice, we just need some paths which carry most of the power from light sources to the points on the camera film and probabilistically terminate the computation. Each integral in the above equation in itself is solved through Monte Carlo integration with sampling probabilities such rays, which will have more light contribution, are samples with high probabilities. 

On a high level, the rendering involves following steps:
\begin{itemize}
    \item Sample rays on the camera film based on the camera model
\item Sample points on the objects in the scene using some probability distribution
\item Try to connect the object point to the light source
\item Collect the light contribution of that path multiplied by the probability of that path
\item Probabilistically terminate paths after certain max length based on some criteria
\end{itemize}
In our system, we use bidirectional path tracing\cite{lafortune1993bi}, which involves paths starting both from camera and light sources. This method is useful for rendering scenes in which light sources are behind translucent objects as in our sensors. 

\subsection{Light models}\label{light_model}
\begin{figure}
\centering
\includegraphics[width=0.8\columnwidth]{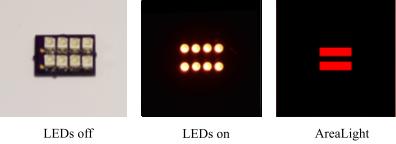}
\caption{\textit{Light model comparison}: The mesh model of the AreaLight model was chosen to match the real LEDs array set as shown in left. Our simulation model matches closely in terms of spatially varying illumination obtained on the sensor surface.}
\label{fig:light}
\end{figure}

We use \textit{AreaLight} model for our simulation system. The \textit{AreaLight} is a good approximation of diffuse illumination received on the deformable surface of our sensor. This light model is fast to simulate and is a common choice for simulating natural lighting \cite{loubet2019reparameterizing} in computer graphics. The key parameters in \textit{AreaLight} are mesh, defining the geometry of the light and 3-dimensional vector for radiance of each color. Figure \ref{fig:light} shows the comparison of physical LEDs used in our prototype sensor and \textit{AreaLight} model. We use differentiable rendering available in Mitsuba2 \cite{nimier2019mitsuba} for obtaining the color intensity for each \textit{AreaLight} set used in our simulation. Please refer to appendix for the details of the optimization procedure. The final optimized values were [5.23, 0.00, 0.00], [0.17,6.73,0.00] and [0.00,0.00,6.83] for red, green and blue LEDs respectively.   

\subsection{Gelcore model}\label{gelcore_model}
\begin{figure}
\centering
\includegraphics[width=\columnwidth]{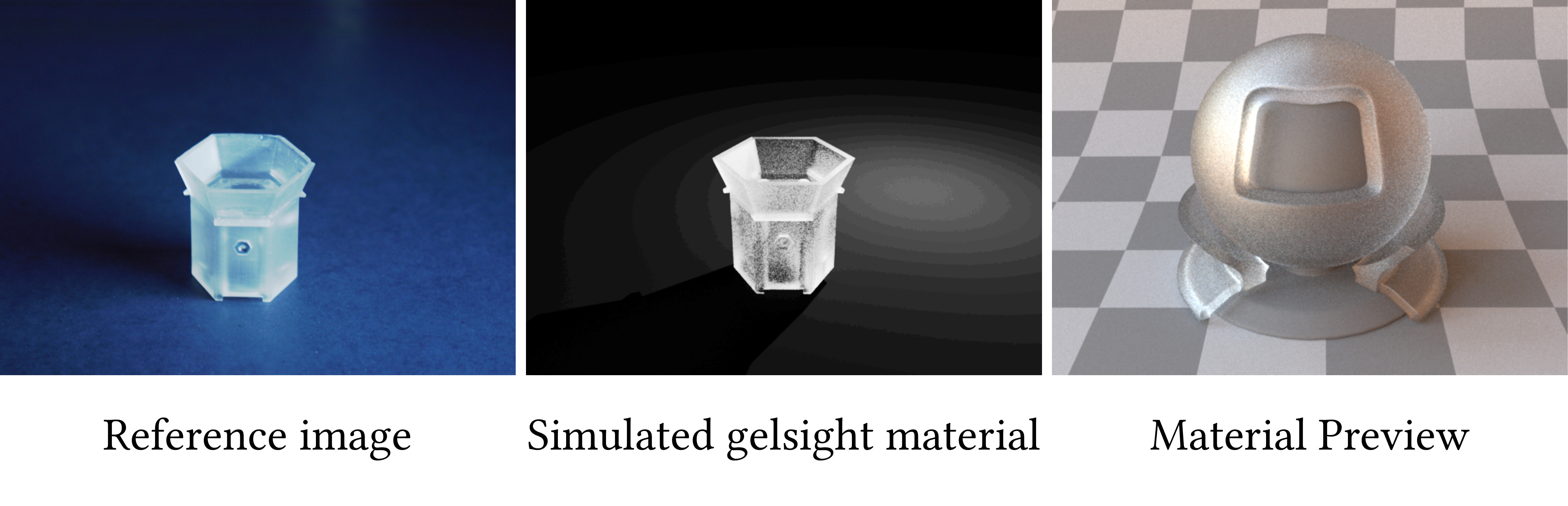}
\caption{\textit{Gelcore material}: The real translucent gelcore in GelSight(left), the simulated Gelcore(middle) with rough dielectric material model and preview of a sphere(right).}
\label{gelcore}
\end{figure}

\textit{Gelcore} refers to the translucent supporting structure inside the sensor, as is visualized in Figure \ref{gelcore}.
The geometric model of the gelcore is exported from the SolidWorks, a 3D geometry modeling tool. The gelcore material is modeled as a dielectric with roughness. The dielectric material model uses microfacet theory\cite{walter2007microfacet} with normals chosen using ggx distribution. The model uses physically accurate fresnel diffraction term, which is essential for modeling scattering losses and roughness. Figure \ref{gelcore} shows the comparison between real gelcore, rendered gelcore and the material visualization using a spherical ball.  
For more complex sensor design materials, one can model the material as a linear combination of different microfacet BRDFs using the isotropic GGX parametric model and its parameter can be optimized using differentiable rendering, as shown in \cite{shem2020towards}.

\subsection{Elastomer surface}\label{elastomer}
In this section, we define the geometric and material model of the soft deformable elastomer surface. We use diffuse material model for the elastomer surface to match the nature of the material's reflectance. This material model is parameterized by a 3-dimensional vector, which describes the ratio of light reflected to that of light received at the surface, respectively. Similar to the light intensities, we used differentiable rendering for optimizing material parameters. After optimization, we obtained [0.50,0.39,0.45] and [0.51, 0.28, 0.22] for flat gel surface and dome-shaped gel surface, which are then used for rendering the corresponding images.  Please refer to appendix for the details of the optimization procedure. 

The 3D geometry of the deformable layer is modeled as a heightfield\cite{tevs2008maximum}. A heightfield is a 2D matrix where each entry represents height at that row and column location. This representation allows to modify the geometry by performing 2D image processing operations on the matrix. 

To obtain the  deformed elastomer surface when an object is pressed against the sensor, we subtract the height of the object from the height of the undeformed elastomer surface.
Note that due to the continuity of the elastomer material, the deformation of the elastomer is a `smoothed-out' shape of the object in contact. We propose a simplified model of this `smoothing out' effect by convolving the object's geometry with a kernel to generate the heightfield of the elastomer surface. The kernel is defined as  

\begin{align}
k(x,y) &= \frac{m+1}{m + \exp{(r\times p)}}\\
\text{where } r &= \sqrt{x^2+y^2}, x,y \in [-\left \lceil{\frac{6}{p}} \right \rceil,\left \lceil{\frac{6}{p}} \right \rceil]
\end{align}

\begin{figure}
\centering
\includegraphics[width=0.5\columnwidth]{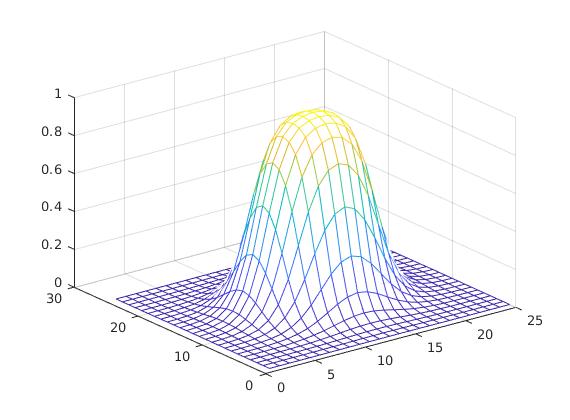}
\caption{Deformation kernel with p=1 and m=200 used to smooth the heightfield}
\label{kernel}
\end{figure}

\begin{figure}
    \centering
    \includegraphics[width=0.8\columnwidth]{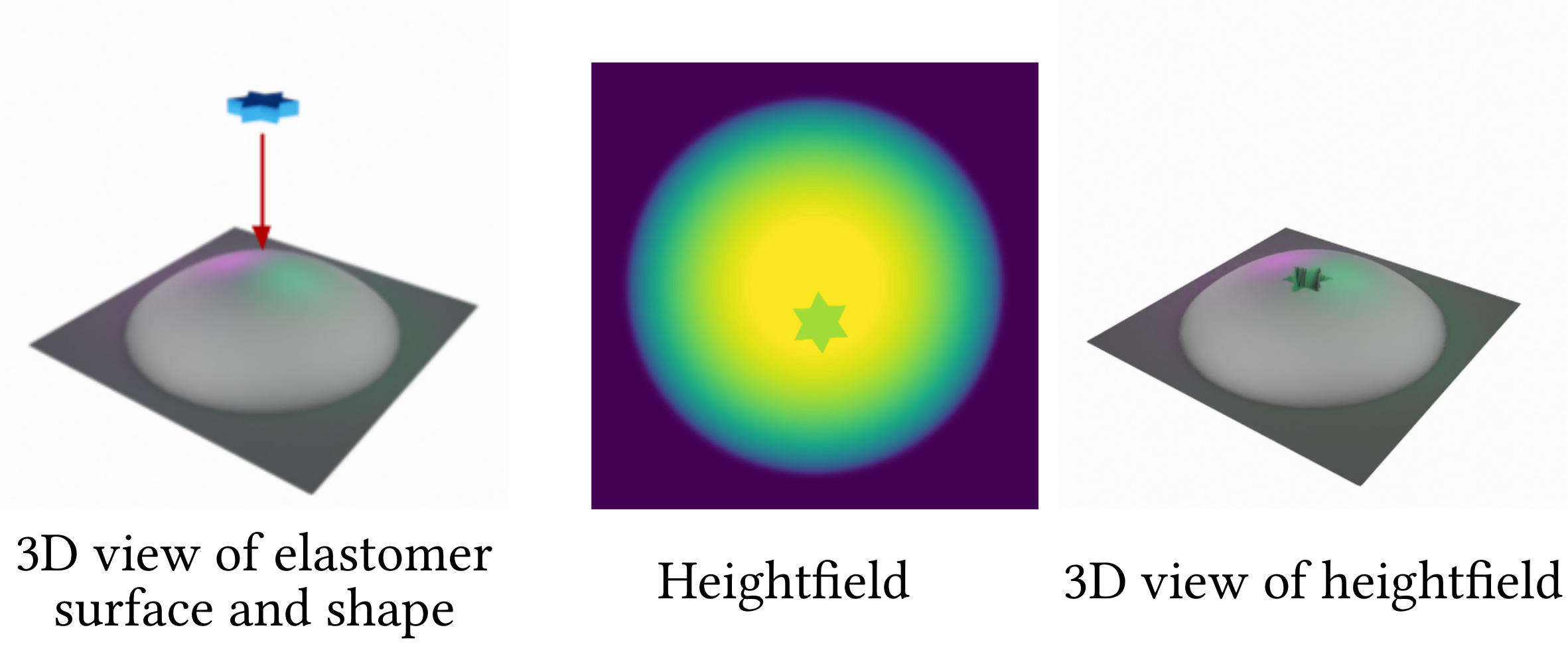}
    \caption{\textit{Heightfield}: Interaction of a star pressed against a sensor (left) and the deformed elastomer surface can be represented using a heightfield image (middle) and the corresponding 3D view is shown on the right.}
    \label{fig:hf}
\end{figure}

This is a simplified method to get material deformation around the edges. However, it is not exact and depends on the depth of press against the sensor. 
For our datasets, we found $p=1$ and $m=200$ works well by hand-tuning the values such that edges of sharp objects match well. Figure \ref{fig:hf} visualizes the 3D view of the undeformed heightfield and deformed heightfield after convolution.

\section{Results and Discussion}

In this section, (\ref{data_collection}) we describe the data collection process and (\ref{light_eval}) the experiments to validate models of light. We then show in (\ref{sensor_eval}) the  sensor simulation when objects of various shapes contact the sensor at various locations.
We used Mitsuba \cite{Mitsuba}, which is an open-source forward renderer with a rich library of material models and light transport integrator methods, for generating GelSight images with models proposed in the paper. The key time-consuming raytracing components are implemented in C++ with GPU acceleration in Mitsuba renderer. We used Mitsuba's python API for rendering all the images. The code to reproduce the experiments is available on \hyperlink{https://github.com/CMURoboTouch/tactile_optical_simulation}{github}.

\subsection{Real Sensor Data Collection}\label{data_collection}
We constructed an optical benchtop setup with 3 translational degrees of freedom.
We mounted the prototype GelSight sensor on a XY movable stage using custom designed 3D printed parts. We mounted the objects to be pressed against the sensor on a vertically movable stage to control depth of press. Our experimental setup is shown in Figure \ref{teaser}. The bench-top setup allows to precisely control the depth of press against the sensor surface and allows to make static indentations. In the GelSight prototype, we used Raspberry Pi V1 camera, as it is compact and allows access to raw images and jpeg images. We plugged the LEDs to a breadboard which allowed us to control the color and intensity of LEDs. 

We collected 2 datasets of real images using our sensor setup. The first dataset contains variation along elastomer surface geometry and indentation depth. The second dataset has variation on the location of contact on the elastomer surface. 

We used a 4mm diameter metal ball and two 3D-printed shapes, namely triangle and star, for pressing against the sensor. The triangle shape offers sharp edges and star shape has multiple interreflection lead to multiple edge colors. The mesh files for the 3D-printed shapes is present on our paper webpage. The first dataset contains 48 images in total with 24 images using flat elastomer surface (3 shapes, 2 pressing heights, 4 elastomer surface locations) and 24 dome elastomer surface (3 shapes, 2 pressing heights, 4 elastomer surface locations). Dome shaped elastomer surface was found to have better light distribution and contact for tactile sensing\cite{gsSensors}. This dataset contains challenging simulation scenarios due to interreflection in star shape, sharp edges in triangle and star, and unknown boundary in metal ball. The second dataset contains 16 images with flat gel elastomer surface in total,  with 10 locations with ball and 6 locations with triangle. This dataset is used to evaluate if the simulation is able to model the variation in light intensity and color at different locations on the elastomer surface. 

\subsection{Lighting model}\label{light_eval}
\begin{figure}[h]
\centering
\includegraphics[width=0.8\columnwidth]{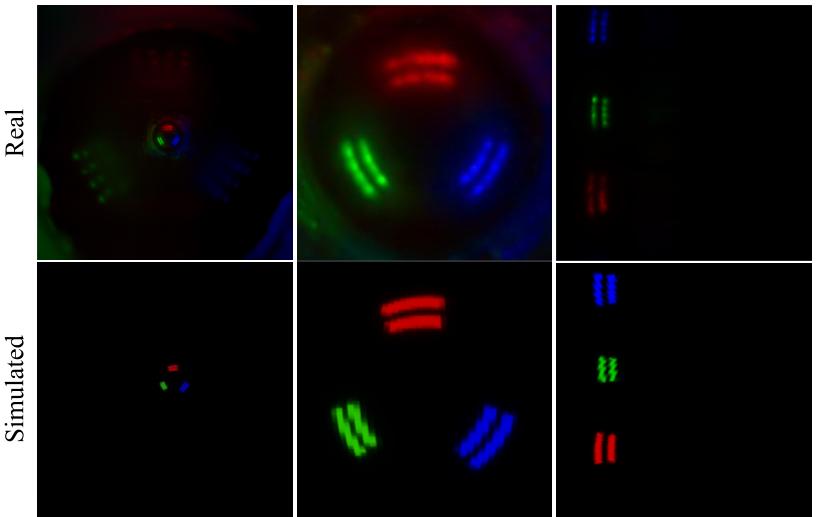}
\caption{\textit{Light probe comparisons}: First column shows the image from the camera viewpoint. Second column compares cropped probe images seen from the camera in real and simulated case. Last column compares the environment map\cite{debevec2000acquiring} of the corresponding image. This image uses our light model without gelcore with optimized scene parameters. The image shows close match of simulated and real world light patterns in terms of shape and color.
}
\label{probecompare}
\end{figure}

In this section, we evaluate the proposed light model and compare its intensity and color at different locations by capturing light probe images\cite{debevec2000acquiring}. The \textit{light probe} refers to a chrome metal ball placed at the location where the image has to be simulated. The light probe image essentially means to capture an image of the scene with only scene lights(sensor LEDs in our case) and the light probe placed at the location where the model has to be tested. 
To capture light probe images from the real sensor, we removed the elastomer surface, and then inserted a metal ball at the same distance as the elastomer surface.  
Figure \ref{probecompare} shows a comparison between real and simulated images for full resolution camera images, cropped light probe images and the corresponding environment map(which represents the light received by the metal ball in polar coordinates). The light probe images show close match in shape of real and simulated light models.

\begin{figure*}[!t]
\centering
\includegraphics[width=\textwidth]{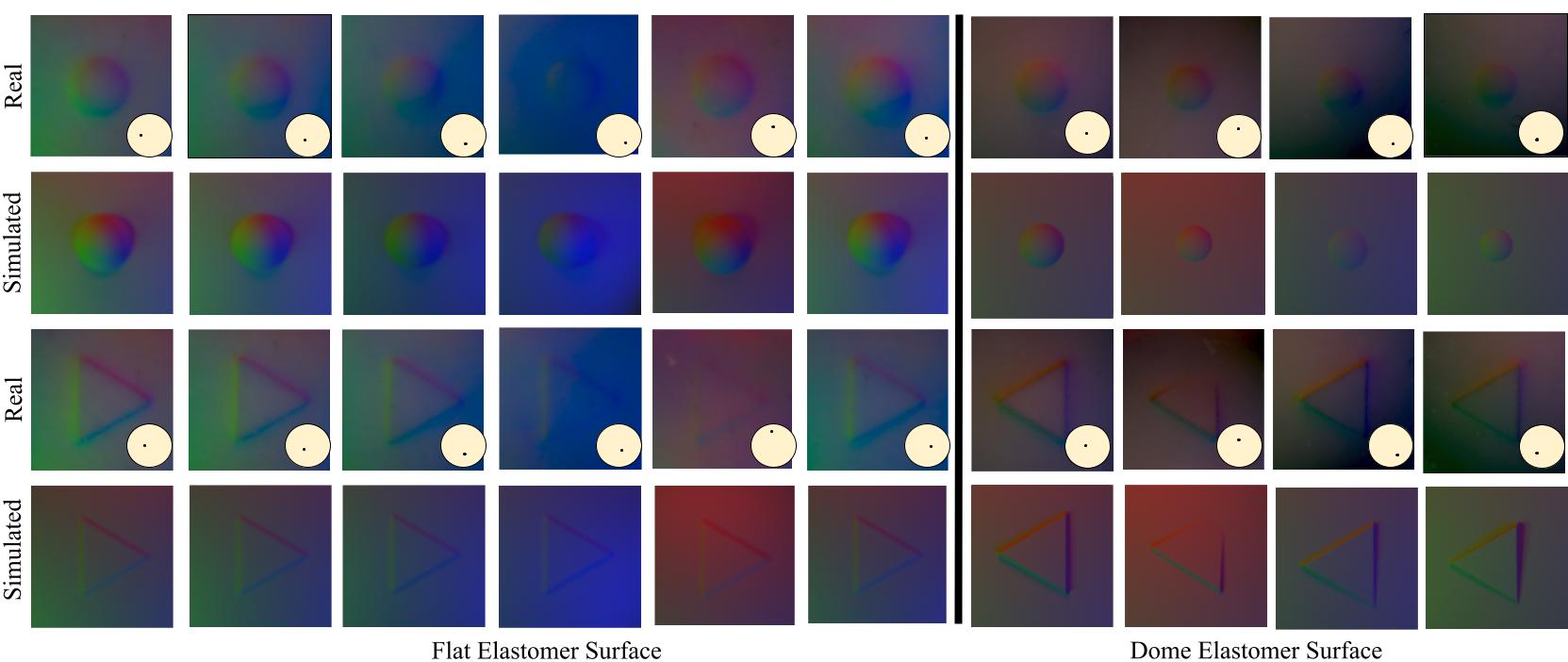}
\caption{\textit{Spatial variation of lighting}: The images in the odd row show 600 $\times$ 600 cropped view of the real sensor images and small inset in the bottom right corner shows where the indentation was made on the original sensor. In all cases the indentation depth was 1mm. The even row shows images rendered using our system. The comparison shows close match in terms of color and intensity of the lighting variation at different parts of the elastomer surface using our simulation system.}
\label{contactloc}
\end{figure*}

\subsection{Evaluation of Sensor simulation}\label{sensor_eval}
In this section, we bring together the light model, gelcore material model and the elastomer surface model to simulate the GelSight sensor when different objects are pressed against the sensor. In our system, we used perspective camera model, with focal length obtained using camera calibration and obtaining camera center using mechanical design. 

We compare against 2 previous approaches, \cite{roberto_shaperecon} and \cite{gomes2019gelsightsim}, which used directional light and phong material for elastomer surface to simulate GelSight; \cite{gsSensors} used directional light assumption and diffuse material for elastomer surface to reconstruct the shape. We use Monte Carlo simulation for both methods as it gives physically accurate results for material models used in above approaches. 

To find the light intensity of our simulation and comparative methods, we collected images with a single color LEDs switched on in our real GelSight prototype and used inverse rendering. For elastomer reflectance color parameter, we used a single image with metal ball pressed in the middle of the sensor and used inverse rendering. We used the mean squared error between the RGB images to estimate the parameters. The details of the optimization procedure could be find in the supplementary material available at our website. 
\textit{Note}: In all our final comparisons, we used sRGB images for visualization and linear images for quantitative evaluations. sRGB images represent the true radiance received by the sensor for each color channel. sRGB images represent images which are post processed for human visualization. We visualize 600$\times$600(which corresponds to 7.1mm $\times$ 7.1mm region in world coordinates) cropped region around the indented shape to show the variation in sensor image.  \textbf{Note}: Previous approaches only show the low resolution image of the full sensor view.

\textbf{Spatial variation}: For this experiment, we used dataset 2, which consists of shapes pressed against sensors at multiple locations. As seen in Figure \ref{contactloc}, our simulation closely matches the colors and intensities for smooth ball and sharp triangles at various locations shown in the inset of the reference images, for multiple elastomer geometry.

\textbf{Rendering complex objects}
\begin{figure}[h]
    \centering
    \includegraphics[width=\columnwidth]{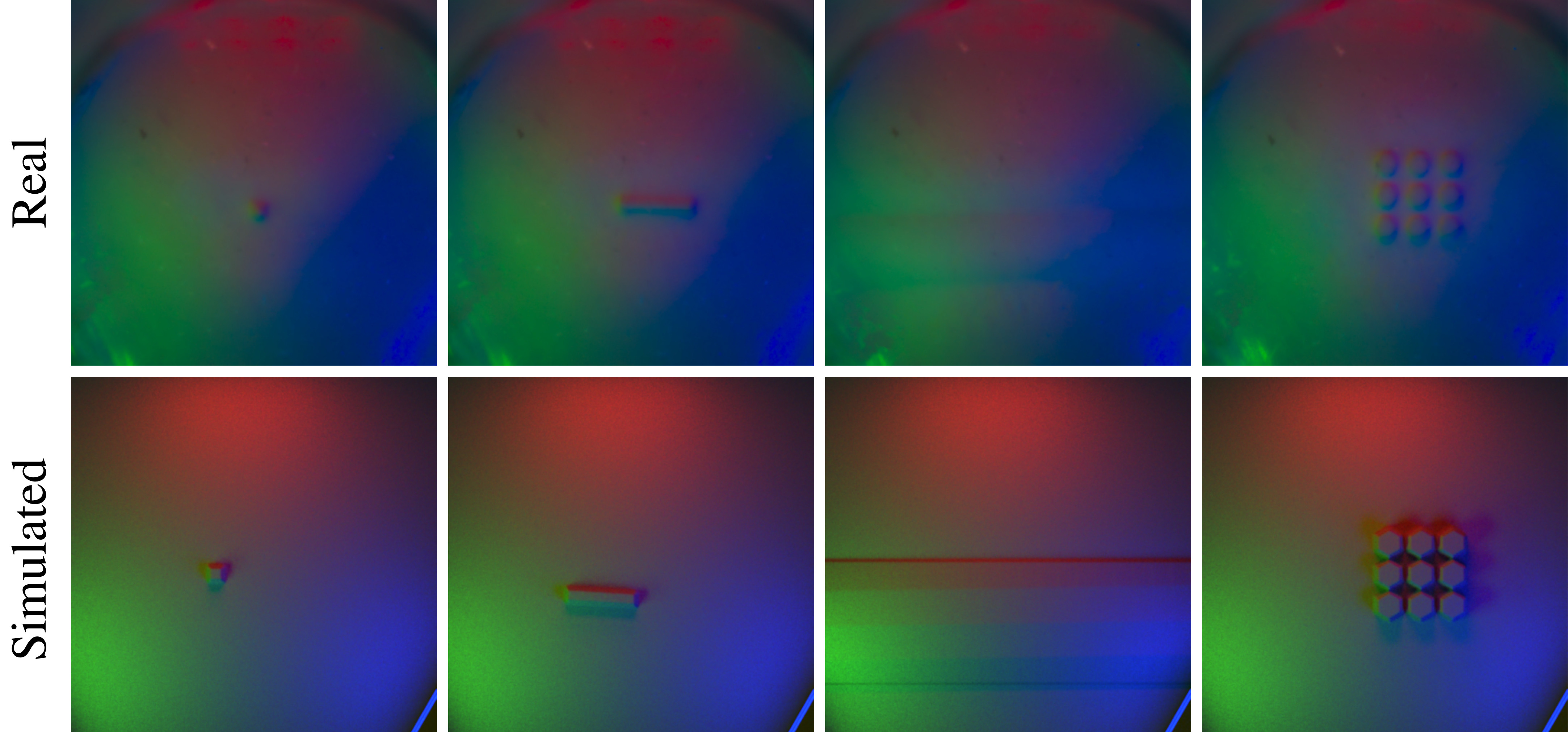}
    \caption{Multiple 3D shapes : Comparison of simulated and real tactile images. Real images were collected by pressing the objects 1mm against the sensor surface }
    \label{fig:complex}
\end{figure}

Figure \ref{fig:complex} shows the qualitative comparison of larger and complex 3D shapes simulated using our system. 

\begin{figure}[h]
\centering
\includegraphics[width=\columnwidth]{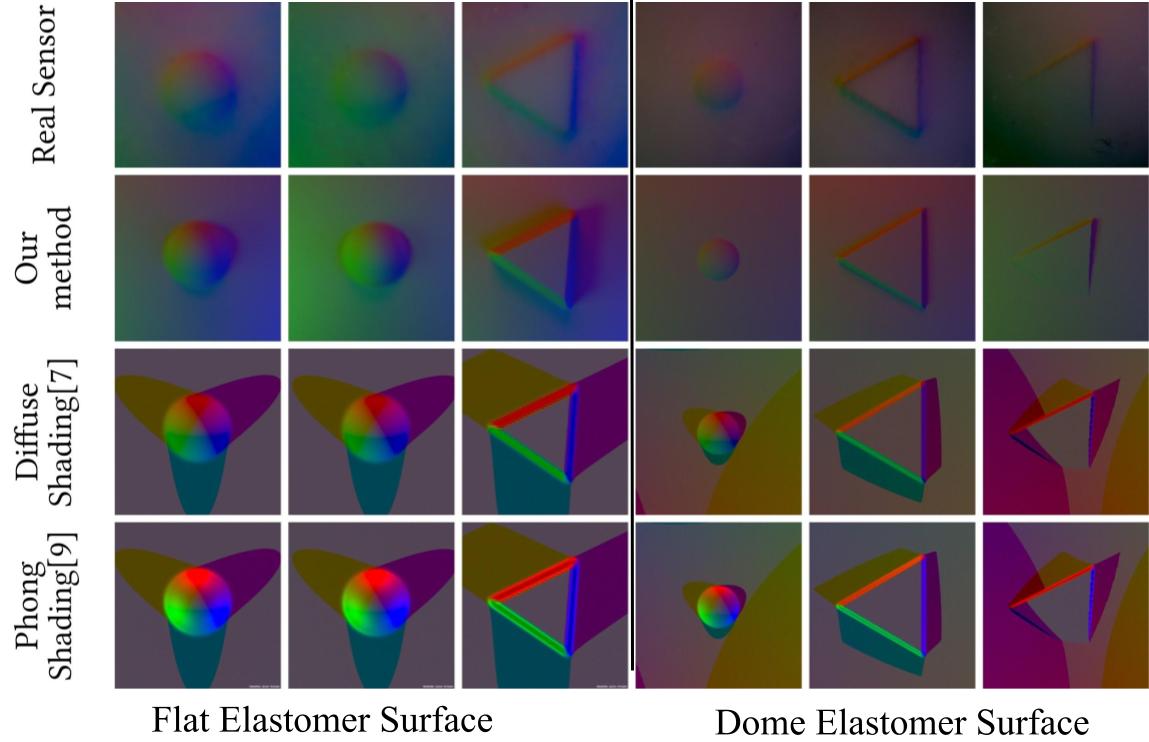}
\caption{\textit{Qualitative comparison between different simulation methods}: The baseline methods are able to capture color and intensity around the center region. However, only our method is able to capture the spatial variation and matches well with the real sensor image for multiple object geometry and elastomer surface geometry.}
\label{methodcompare}
\end{figure}

\textbf{Comparison against other methods}: For this experiment, we used dataset 1, which consists variations along shapes, indentation depth and elastomer surface geometry. For qualitative comparison refer to Figure \ref{methodcompare}, we show balls pressed at different locations in the first 2 columns. Our method closely matches the strong green color in the 2th column. Directional light model assumes that the light intensity remains constant across the elastomer surface. This affect is seen in column 1-2, where the image remains same irrespective of where the ball was pressed on the sensor surface. In the 3th column only our method shows correct colors at the edges of triangle.  Column 4-6 show the comparison of shapes pressed against a dome shaped elastomer surface. Dome shape is particularly challenging due to concave shape of the elastomer surface, which can have strong interreflections. As can be seen, our method has correct colors at the edges and shows variation in the color when shapes are pressed at different locations at the sensor. In real sensor the intensity of light increases if the location of contact is closer to a light source. However, previous method assume constant lights and fail to accurately capture the color at locations close to light sources. 

For quantitative comparison, we used traditionally signal processing metrics like Mean Squared Error(MSE), Signal-To-Noise Ratio(SNR), Symmetric mean absolute percentage error (SMAPE) and a metric from image similarity literature SSIM\cite{hore2010image}. SSIM is insensitive to  luminance change, contrast change and small geometric distortions. It produces a single number per-pixel by finding the mean, variance and correlation per channel. To compare images, one can take the mean over all the pixels to obtain a single number for the image known as Mean SSIM, which we use in this paper. We used cropped patch of 600 $\times$ 600 around the indented shape for calculating metrics. Table \ref{imgsim} shows the average metric values calculated using images from both the datasets. We consistently outperform previous methods on a range of image similarity metrics.


\begin{table}[h]
\caption{Comparison of the simulated tactile images and the real ones on different metrics. The $\downarrow$ arrow shows that lower value is desired and vice-versa.}\label{imgsim}
\centering
\begin{tabular}{|p{2.2cm}|c|c|c|c|}
\hline
& MSE $\downarrow$ & SNR $\uparrow$ & SMAPE $\downarrow$ & SSIM $\uparrow$ \\
\hline
Diffuse surface + Directional Light\cite{gsSensors} & 0.003807 & 1.46 & 0.84 & 0.42  \\
Phong shading + Directional Light\cite{gomes2019gelsightsim} & 477.44 & -35.39 & 0.87 & 0.40  \\
\hline
\textbf{Our method} & \textbf{0.001710} & \textbf{6.72} & \textbf{0.47} & \textbf{0.80}  \\
\hline
\end{tabular}
\end{table}

\section{Conclusion}
In this paper, we introduce the first fully general optical simulation system for vision-based tactile sensors. We use physics based light transport simulation with physically accurate light models, gelcore material and  elastomer surface representation. We verify our physically valid models for lights, material and geometry by using a bench-top setup and a GelSight prototype. Our results consistently highlight the benefits of using physically grounded optical simulation through quantitative and qualitative comparison with multiple sensors and object variations. Our system is able to model the variation in light intensity and color in presence of challenging inter-reflections, sharp edges and multiple surface geometry. 

The one downside of our method is that we currently process the mesh by intersecting surface and object offline. This process can be made trivially parallel as we use convolution, which is highly optimized on GPU. Another major challenge in our system is that GPU has limited memory. Therefore simulating large scenes could be challenging. Another key challenge is running the simulation at interactive frame-rates. Due to advancements in both hardware\cite{nvturing} and software\cite{gharbi2019sample}, realtime raytracing is being actively used in films\cite{christensen2018renderman} and games. Therefore, we anticipate that this problem would be solved in the near future.  



\bibliographystyle{IEEEtran}
\bibliography{ref}

\clearpage
\appendix
\subsection{Ablation study : Per-channel image comparison}
\begin{figure}[h]
\centering
\includegraphics[width=\columnwidth]{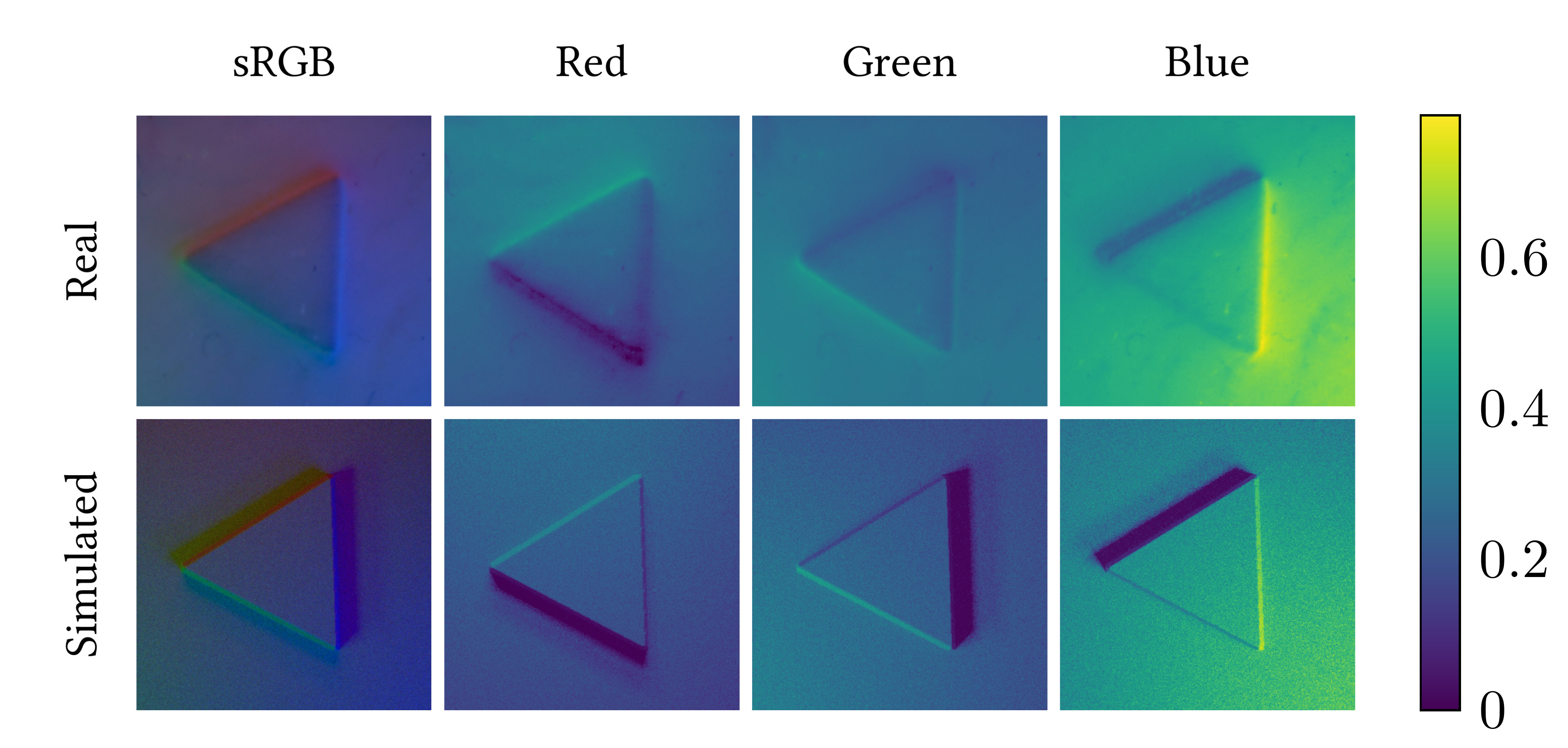}
\caption{Comparison between real and simulated image in different color channels}
\label{colorcomp}
\end{figure}

GelSight uses individual color channel information for shape reconstruction. Therefore, individual color channels are important to inspect. For this reason, we considered a case when a triangle is pressed against flat elastomer surface at a depth of 1mm. Figure \ref{colorcomp} shows the comparison of RGB channels between the real and the simulated sensor images. The figure shows close match in terms of light intensity in all the channels especially high values on the right side in Blue channel. Though, we note that the simulated images have large shadows which are missing in real images. However, the edges of shapes match closely in real and simulated images.

\subsection{Timing and image quality trade-off}
\begin{figure}
    \centering
    \includegraphics[width=\columnwidth]{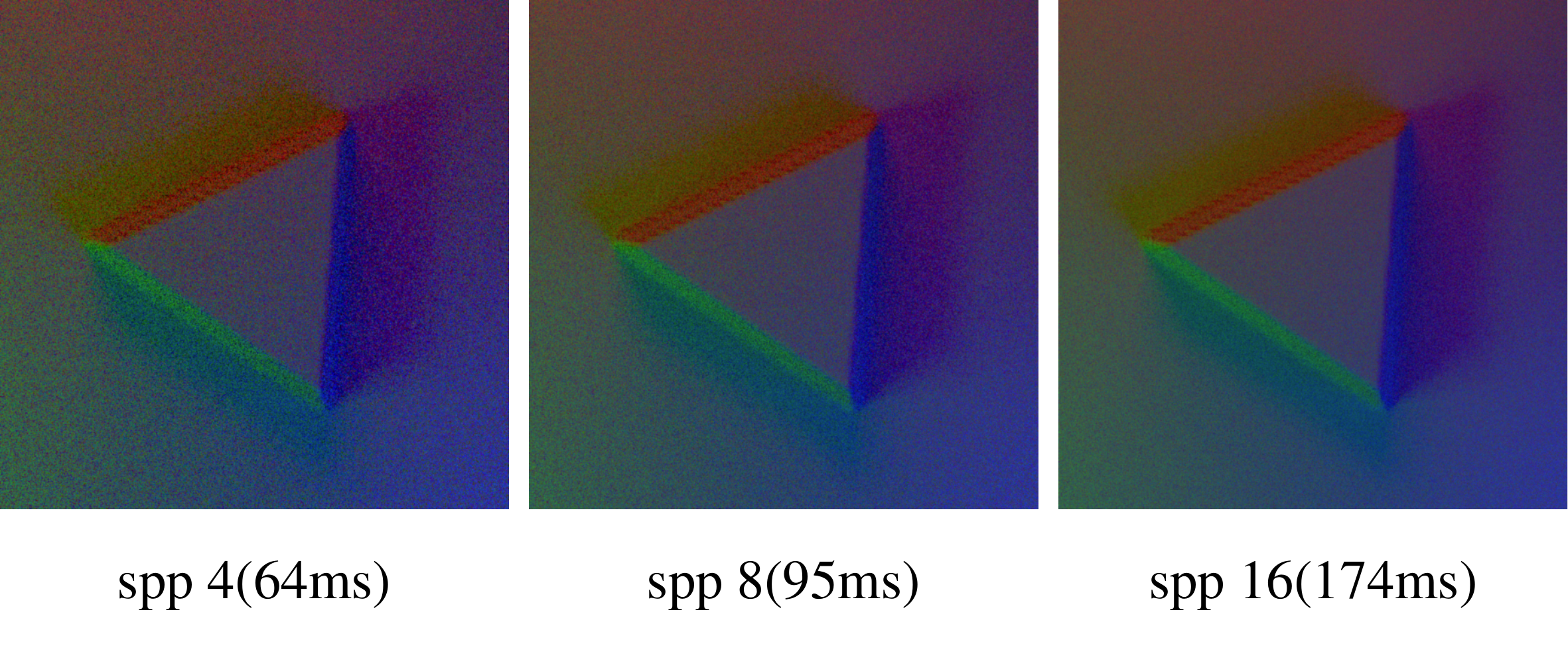}
    \caption{Qualitative image comparison of images rendered at different sample per pixel(SPP)}
    \label{fig:spp}
\end{figure}

We use Monte Carlo integration in our simulation which uses sampling to obtain the color of a pixel. Therefore the total time taken to render each image and its quality is a function of sample-per-pixel(spp), length of light path($l$) and size of the image($h \times w$).  The quality of the rendered image and computation time both increase if we increase any of the mentioned parameters. We found in our experiments $spp=8, l=4$ for rendering image size 600 x 600 leads to frame-rates of 10Hz and is optimal. While keeping image size and $l$ constant, the timings for spp 4, 8 and 16 are 64ms, 95ms and 174ms. Figure \ref{fig:spp} shows the image comparison with varying spp. While keeping spp and $l$ constant, the timings for image size 128x128, 256x256, 512x512 and 1024x1024 are 33ms, 36ms, 55ms and 126ms. These runtimes were recorded using python3.7 $timeit$ module on 32 CPU core machine with GeForce RTX 2080Ti for a scene with 318510 geometric faces. Although, we perform timing evaluation, the actual simulation speed depends on the complexity of the scene. 

\subsection{SSIM map comparison}
\begin{figure*}[!t]
\centering
\includegraphics[width=\textwidth]{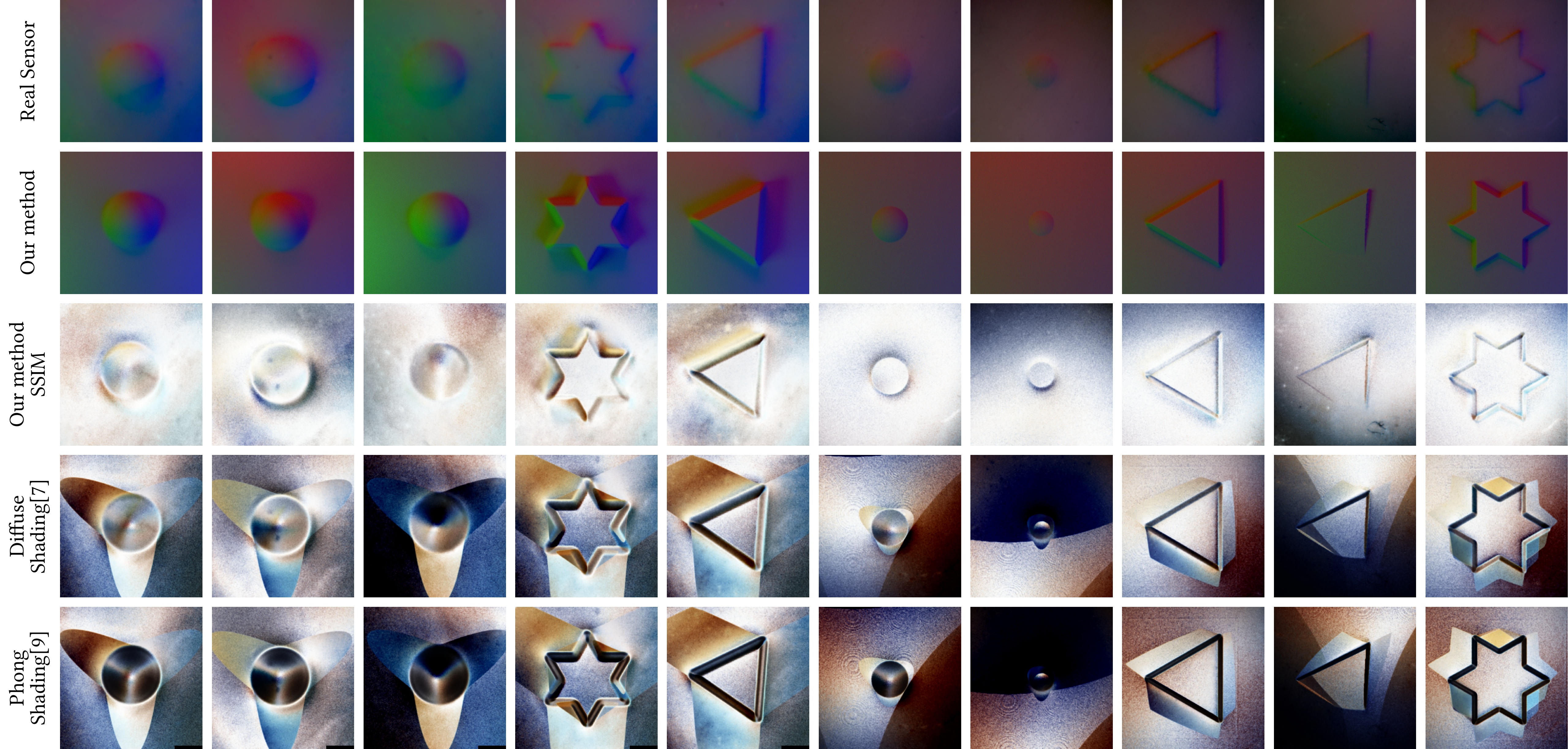}
\caption{\textit{SSIM map comparison between different simulation methods}: The baseline methods are able to capture color and intensity around the center region. However, only our method is able to capture the spatial variation and matches well with the real sensor image for multiple object geometry and elastomer surface geometry.}
\label{ssimmethodcompare}
\end{figure*}

Structural Similarity Index \cite{hore2010image} provides per pixel similarity map between simulated and real image, which can be used to analyze specific image regions. We calculate SSIM map per channel, which differs from original formulation, that uses luminance channel for a colored image. We compare the SSIM maps for our method and compare against other methods. SSIM maps should be a white image is the test image if close to reference image. Figure \ref{ssimmethodcompare} shows the SSIM maps.

\subsection{Model parameter optimization}
We used a two-stage approach for obtaining the parameters of our simulation system. We optimize for 12 parameters in total : 3 for intensity of each light source and 3 for surface reflectance. We optimize for flat elastomer gelpad and dome shaped elastomer gelpad separately. 

\begin{figure}[!h]
    \centering
    \includegraphics[width=\columnwidth]{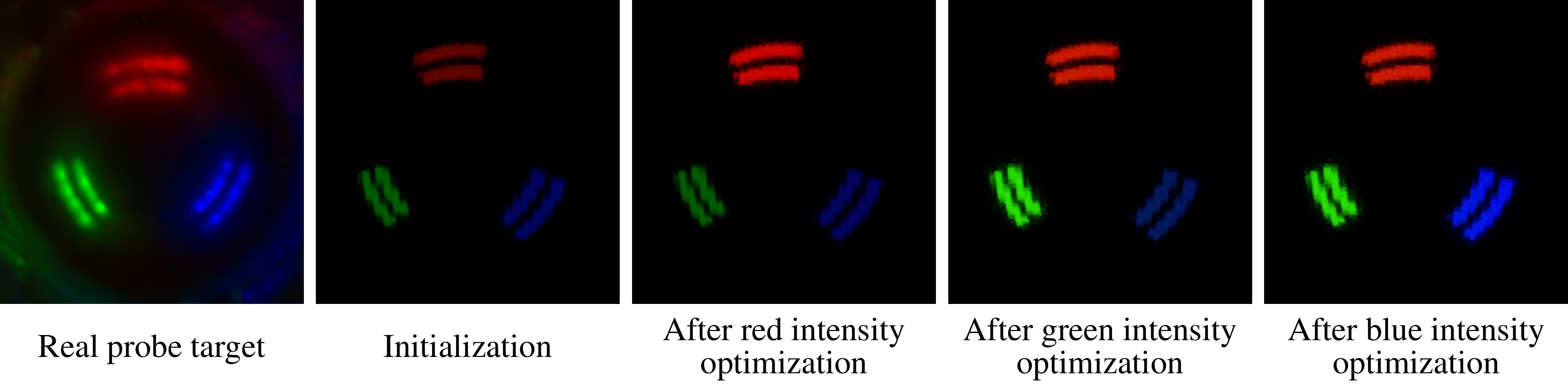}
    \caption{Visualization of the light probe at different stages of intensity optimization}
    \label{fig:optim1}
\end{figure}

In the first stage, we used a light probe image to optimize for 3-dimensional intensity vector for each \textit{AreaLight} separately. The order of optimization was red, green and blue. Figure \ref{fig:optim1} shows the images at different optimization steps. Note we are only concerned about the regions which are directly visible near the light pattern, as we use the cropped region around the bright regions for optimization. The final optimized values for red, green and blue LEDs are [5.23, 0.00, 0.00], [0.17,6.73,0.00] and [0.00,0.00,6.83]

\begin{figure*}[!h]
    \centering
    \includegraphics[width=\linewidth]{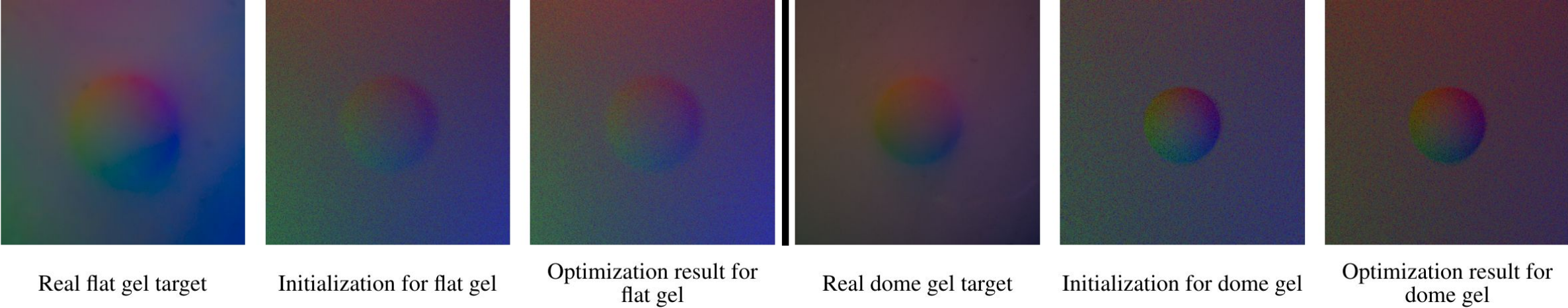}
    \caption{\textit{Visualization of optimization procedure for surface reflectance}: The initialization starts with uniform reflectance(1,1,1) and then a differentiable rendering optimization procedure is used to minimize the loss for the specific gelpad surface geometry. The first 3 images on the left show the target, initialization and final image after the optimization procedure for the flat gelpad surface geometry. The next 3 images after the separating line show the corresponding images for the domegel gelpad surface geometry. Note: domegel surface is noted to be better for shape reconstruction.}
    \label{fig:optim2}
\end{figure*}

The second stage of the parameter optimization tries to recover a 3-dimensional surface reflectance property of the deformable elastomer surface of the sensor. We use a single image with sphere pressed against the sensor for both the gelpads. Figure \ref{fig:optim2} shows the visualization of optimization for both flat gelpad and dome gelpad. The optimized surface reflectance for flatgel and dome gel surfaces are [0.50,0.39,0.45] and [0.51, 0.28, 0.22] respectively. 

\subsection{3D Printed shapes}
\begin{figure}[!t]
\centering
\includegraphics[width=0.8\columnwidth]{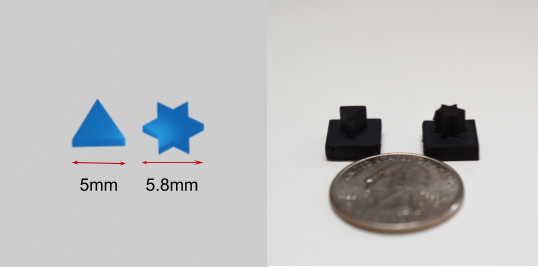}
\caption{\textit{3D printed shapes}: The image on the left shows the shapes visualized in blender and image on the right shows the real shapes placed beside a US quarter. These shapes are pressed against our sensor for data collection.}
\label{3dprinted}
\end{figure}

Figure \ref{3dprinted} visualizes the shapes used in our experiments. Please visit our paper website for mesh files, which can be directly used for 3D printing the shapes. 

\subsection{Low-level processing in dataset collection}
We collected jpeg + raw images from raspberry pi using picamera library. We used dcraw to extract raw images. The raw images are then undistorted and stored in exr format using matlab.  
The collected images were hand-annotated for finding the object location w.r.t to sensor surface. We used the camera parameters to obtain world coordinates of the objects. The world coordinates of the objects were 
used to generate heightfields and place the generated geometry into the simulation environment. 
\end{document}